\documentclass[letterpaper,conference]{IEEEtran}
\usepackage[utf8]{inputenc}
\usepackage[T1]{fontenc}
\newcommand{\mR}{\mathbb{R}}
\newcommand{\mnorm}[1]{\left\lVert#1\right\rVert}
\newcommand*{\defeq}{\mathrel{\vcenter{\baselineskip0.5ex \lineskiplimit0pt
                     \hbox{\scriptsize.}\hbox{\scriptsize.}}}%
                     =}
\newcommand{\comment}[1]{}
\usepackage{bm}
\usepackage{amsmath}
\usepackage{amsfonts}
\usepackage{amsthm}
\usepackage{graphicx}
\usepackage{cite}
\usepackage{url}
\usepackage{booktabs}
\usepackage{tablefootnote}
\usepackage{subcaption}

\renewcommand{\comment}[1]{\ignorespaces}
\begin{document}
\title{On Model Predictive Path Following and Trajectory Tracking
  for Industrial Robots} 
\author{\IEEEauthorblockN{Mathias Hauan
    Arbo\IEEEauthorrefmark{1}, Esten Ingar
    Gr{\o}tli\IEEEauthorrefmark{2} and Jan Tommy
    Gravdahl\IEEEauthorrefmark{1}}
  \IEEEauthorblockA{\IEEEauthorrefmark{1}Department of Engineering Cybernetics\\
    NTNU, Norwegian University of Science and Technology}
  \IEEEauthorblockA{\IEEEauthorrefmark{2}Mathematics and Cybernetics,
    SINTEF DIGITAL, Trondheim, Norway}}

\maketitle
\begin{abstract}
  In this article we show how the model predictive path following
  controller allows robotic manipulators to stop at obstructions in a
  way that model predictive trajectory tracking controllers cannot. We
  present both controllers as applied to robotic manipulators,
  simulations for a two-link manipulator using an interior point
  solver, consider discretization of the optimal control problem using
  collocation or Runge-Kutta, and discuss the real-time viability of
  our implementation of the model predictive path following controller.
\end{abstract}

\section{Introduction}
%\subsection{Motivation}
Modern industrial robots weld, grind, screw, measure, film, paint,
pick and place, and perform other tasks that require the robot to
follow some geometric path in space. For a typical robot cell, we
simplify our work as robotics engineers by having enclosed, structured
workspaces with no obstructions. For dedicated large-scale production
of a small set of products, this is easily achieved. For small-scale
manufacturing however, there is a large variety of products and each
product series is produced in low-scale or on-demand. This calls for
rapid prototyping of both the paths the robot has to move, and its
environment. We must consider control strategies that can handle both
unknown and known obstructions without sacrificing quality of the
product.

We consider model predictive control (MPC), where we may define
constraints on states, and rudimentary obstructions. This article
considers known obstructions, and focuses on the difference between
path-following and trajectory tracking.

In trajectory tracking model predictive control (TT-MPC), the robot is
to follow a path with an explicit path-timing. The trajectory may even
incorporate constraints on the torques or velocities of the robot
through optimal control problem (OCP) approaches such as
\cite{Verscheure2009}, where a time-optimal path timing law under
constraints is generated. The path timing law specifies the relation
between the desired path and time, while accounting for state
constraints during the execution of the path. The OCP is solved under
the assumption that the robot is moving along the path, and is an
open-loop approach to the constraint handling. In
\cite{Debrouwere2013} this was extended to also allow constraints on
the acceleration and inertial forces at the end-effector.

In~\cite{Faulwasser2009}, a model predictive path-following controller
(MPFC) that handles both path-timing and error from path in the same
OCP is described. In~\cite{Faulwasser2016conv}, the MPFC is shown to
converge to the path given terminal constraints without needing
terminal penalties. In~\cite{Faulwasser2016} the MPFC is implemented
on a KUKA LWR IV robot, without end penalty or a terminal
constraint. This is done with the ACADO framework \cite{Houska2011a},
which uses a sequential programming method (SQP), iteratively solving
quadratic programs approximating the nonlinear program using the
qpOASES active set solver.

In~\cite{Lam2010}, a real-time MPFC scheme for contouring control of an
x-y table is described. Here a linear time varying approximation of
the dynamics is used to define a QP which is solved using an
active-set solver. In~\cite{Lam2011}, this is implemented on an x-y
table, and the MPFC outperformed both a similarly implemented TT-MPC,
and the industry standard cascaded PI controlled set-point controller
operating at a higher sampling frequency.

In~\cite{Bock2014}, an MPFC is applied to a tower crane. The
OCP is solved using the gradient projection method, an indirect method
where Pontryagin's Maximum Principle is solved for the OCP without
inequality constraints on the states. Instead slack variables are
introduced to implicitly handle the inequality constraints.

In this article we
\begin{itemize}
\item Draw attention to differences in MPFC and TT-MPC behavior, with and without obstructions,
\item compare the Runge-Kutta and collocation integration method for the two strategies,
\item solve the nonlinear programs (NLP) for the control strategies using the interior point solver IPOPT,
\item and discuss the framework for real-time applications.
\end{itemize}

\section{Theory}\label{sec:theory}
\subsection{System}\label{sec:theory:system}
The robot is an $n$ degrees-of-freedom system with the state-space
representation
\begin{subequations}
  \begin{align}
    &\dot{\bm{y}}(t) = \bm{x}(t)\label{eq:system-y}\\
    &\dot{\bm{x}}(t) = \bm{f_x}(\bm{y}(t),\bm{x}(t),\bm{u}(t))\label{eq:system-x}
  \end{align}
\end{subequations}
where $\bm{y}\in\mR^n$ are the generalized coordinates,
$\bm{x}\in\mR^n$ are generalized velocities, and $\bm{u}\in\mR$ are
the inputs. The function $\bm{f_x}$ describes acceleration. We assume
the robot has known forward kinematics, allowing us to define a point
of interest $\bm{p}$ on the robot such that
\begin{align}
  \bm{p}(t) = \bm{f_{p}}(y(t))
\end{align}
where $\bm{f_p}(\cdot)\in\mR^{n_p}$ is found from the forward
kinematics.  We describe the $\mathcal{C}^1$ reference path as
$\bm{\varrho}(\cdot)\in\mR^{n_p}$, defined in the same frame as
$\bm{p}$.

The path-timing variable $s$ moves from 0 to $s_f$. The TT-MPC assumes
$s=t$ for $t<s_f$ and $s=s_f$ otherwise. The MPFC controls $s$ through
the path-timing dynamics, which we model as a double integrator
\begin{align}
  \begin{bmatrix}
    \dot{s}(t)\\
    \ddot{s}(t)
  \end{bmatrix}=
  \begin{bmatrix}
    0 & 1\\
    0 & 0
  \end{bmatrix}
        \begin{bmatrix}
          s(t)\\
          \dot{s}(t)
        \end{bmatrix}
+
  \begin{bmatrix}
          0\\
          1
        \end{bmatrix}v(t) \defeq \bm{f_s}(s,\dot{s},v),
\end{align}
with piecewise constant input $v(\cdot)\in\mR$. To ensure that we
never move backwards along the path, and that we have a maximum along
path speed, we constrain $\dot{s}\in[0,\dot{s}_u]$. For more
information on choice of the path-timing dynamics, we refer the reader
to \cite{Faulwasser2016conv}.

For the MPFC we define the extended state
$\bm{\xi} = [\bm{y}^T,\bm{x}^T,s,\dot{s}]^T$, input
$\bm{w}=[\bm{u}^T,v]^T$, and dynamics
\begin{equation}
  \dot{\bm{\xi}}(t) =
  \begin{bmatrix}
    \bm{x}(t)\\
    \bm{f_x}(\bm{\xi}(t),\bm{w}(t)\\
    \bm{f_s}(\bm{\xi}(t),\bm{w}(t)
  \end{bmatrix} = \bm{f_{\xi}}(\bm{\xi}(t),\bm{w}(t)).
\end{equation}
Similarly, for the TT-MPC we define the extended state
$\bm{\chi}=[\bm{y}^T,\bm{x}^T]$ with dynamics
\begin{equation}
  \dot{\bm{\chi}}(t) =
  \begin{bmatrix}
    \bm{x}(t)\\
    \bm{f_x}(\bm{\chi}(t),\bm{u}(t)
  \end{bmatrix} = \bm{f_{\chi}}(\bm{\chi}(t),\bm{u}(t)).
\end{equation}

We define the deviation from the path as 
\begin{equation}
  \bm{e}_{pf}(t) \defeq \bm{f_{p}}(\bm{y}(t)) - \bm{\varrho}(s(t)).
\end{equation}
for the MPFC, and 
\begin{equation}
  \bm{e}_{tt}(t) \defeq \bm{f_{p}}(\bm{y}(t)) - \bm{\varrho}(t)
\end{equation}
for the TT-MPC. For $s$ to converge to $s_f$ we also define the
path-timing error
\begin{equation}
  e_s(t) \defeq s(t) - s_f.
\end{equation}

\subsection{Optimal Control Problem}\label{sec:theory:optimal-control-problem}
With the previously defined dynamics and errors, we can describe the
OCP for the MPFC as
\begin{subequations}\label{eq:theory:pf-ocp}
  \begin{align}
    \min_{\bm{e}_{pf},\dot{\bm{e}}_{pf},e_s,\bm{u},v} &\int_{t_k}^{t_k+T} J_{pf}(\tau,\bar{\bm{\xi}}(\tau),\bar{\bm{w}}(\tau))\mathrm{d}\tau\\
    \text{s.t.:}&\nonumber\\
    &\dot{\bar{\bm{\xi}}}(\tau) = \bm{f_{\xi}}(\bar{\bm{\xi}}(\tau),\bar{\bm{w}}(\tau))\label{eq:theory:pf-ocp:state-dynamics}\\
    &\bar{\bm{\xi}}(t_k) = \bm{\xi}(t_k)\\
    &\bar{\bm{\xi}}(\tau)\in[\bar{\bm{\xi}}_l,\bar{\bm{\xi}}_u]\label{eq:theory:pf-ocp:state-constraints}\\
    &\bm{h}_c(\bar{\bm{\xi}}(\tau)) \leq 0 \label{eq:theory:pf-ocp:constraint-function}
  \end{align}
\end{subequations}
and for the TT-MPC as
\begin{subequations}\label{eq:theory:tt-ocp}
  \begin{align}
    \min_{\bm{e}_{tt},\dot{\bm{e}}_{tt},e_s,\bm{u}} &\int_{t_k}^{t_k+T} J_{tt}(\tau,\bar{\bm{\chi}}(\tau),\bar{\bm{u}}(\tau))\mathrm{d}\tau\\
    \text{s.t.:}&\nonumber\\
    &\dot{\bar{\bm{\chi}}}(\tau) = \bm{f_{\chi}}(\bar{\bm{\chi}}(\tau),\bar{\bm{u}}(\tau))\label{eq:theory:tt-ocp:state-dynamics}\\
    &\bar{\bm{\chi}}(t_k) = \bm{\chi}(t_k)\\
    &\bar{\bm{\chi}}(\tau)\in[\bar{\bm{\chi}}_l,\bar{\bm{\chi}}_u]\label{eq:theory:tt-ocp:state-constraints}\\
    &\bm{h}_c(\bar{\bm{\chi}}(\tau)) \leq 0 \label{eq:theory:tt-ocp:constraint-function}
  \end{align}
\end{subequations}
where subscript $u$ refers to the upper bounds on the states,
subscript $l$ refers to the lower bound, and $\bm{h}_c$ describes
other constraints such as obstacles in the path. The bar is to
differentiate internal states of the MPC and the actual system.

The cost integrands are defined as
\begin{align}
  J_{pf}(\tau,\bar{\bm{\xi}}(\tau),\bar{\bm{w}}(\tau)) = &\frac{1}{2}\bar{\bm{e}}_{pf}(\tau)^T\bm{Q}\bar{\bm{e}}_{pf} +\frac{1}{2}\dot{\bar{\bm{e}}}_{pf}(\tau)^T\bm{Q}_d\dot{\bar{\bm{e}}}_{pf}\nonumber\\
                                        &+\frac{1}{2}\bar{\bm{u}}(\tau)^T\bm{R}\bar{\bm{u}}(\tau)\nonumber \\
                                        &+\frac{1}{2}q \bar{e}_s(\tau)^2 + \frac{1}{2}r \bar{v}(\tau)^2
\end{align}
for the MPFC and  
\begin{align}
  J_{tt}(\tau,\bar{\bm{\chi}}(\tau),\bar{\bm{w}}(\tau)) = &\frac{1}{2}\bar{\bm{e}}_{tt}(\tau)^T\bm{Q}\bar{\bm{e}}_{tt} +\frac{1}{2}\dot{\bar{\bm{e}}}_{tt}(\tau)^T\bm{Q}_d\dot{\bar{\bm{e}}}_{tt}\nonumber\\
                                        &+\frac{1}{2}\bar{\bm{u}}(\tau)^T\bm{R}\bar{\bm{u}}(\tau)
\end{align}
for the TT-MPC. The matrices $\bm{Q}$, $\bm{Q}_d$, and $\bm{R}$ are
positive definite. The scalars $q$ and $r$ are positive. We have
included the derivative of the path deviation to reduce oscillations.

Solving the OCPs can be done by an indirect approach using
Pontryagin's Maximum Principle as done in \cite{Bock2014}, or a direct
approach as done in~\cite{Lam2010}. We will apply the direct
simultaneous approach, reformulating the OCP as an NLP by discretizing
the problem. The direct simultaneous approach is most common in
real-time applications, with existing software support such as ACADO
\cite{Houska2011a} and CasADI \cite{Andersson2013b}.

\subsection{Nonlinear Program and Interior Point}
In this section we only give the discretization of
(\ref{eq:theory:pf-ocp}) as the TT-MPC is similar and simpler. To
discretize the OCP we use two different integration methods: the 4th
order Runge-Kutta (RK4), and collocation based on Lagrange polynomials
with $d$ Legendre points.

The control input is a piecewise continuous function, constant on
intervals of length $\delta_t$, which is the length of our
timesteps. This gives us a horizon of length $N_T = T/\delta_t$. With
the simultaneous approach we use the integration method between each
time step, and constrain the result and next state to be equal.

\subsubsection{Runge-Kutta}
Given a $\delta_t$, RK4\cite{modsim} gives us the equation
\begin{align}
  \bar{\bm{\xi}}_{k+1} = \bar{\bm{\xi}}_k + \frac{\delta_t}{6}(k_1+2k_2+2k_3+k_4) = F(\bar{\bm{\xi}}_k,\bar{\bm{w}}_k) 
\end{align}
with
\begin{subequations}
\begin{align}
  &k_1 = \bm{f_{\xi}}(\bar{\bm{\xi}}_k,\bar{\bm{w}}_k),\\
  &k_2 = \bm{f_{\xi}}\left(\bar{\bm{\xi}}_k+k_1\frac{\delta_t}{2},\bar{\bm{w}}_k\right),\\
  &k_3 = \bm{f_{\xi}}\left(\bar{\bm{\xi}}_k+k_2\frac{\delta_t}{2},\bar{\bm{w}}_k\right),\\
  &k_4 = \bm{f_{\xi}}\left(\bar{\bm{\xi}}_k+k_3\delta_t,\bar{\bm{w}}_k\right).
\end{align}
\end{subequations}

The resulting NLP is then
\begin{subequations}\label{eq:theory:nlp}
\begin{align}
  \min_{\bm{q}}&\quad{\phi}(\bm{q})\label{eq:theory:nlp:objective}\\
  \text{s.t.:}&\nonumber\\
  &\bm{f_e}(\bm{q}) = \bm{0}\label{eq:theory:nlp:equality-constraints}\\
  &\bm{h_e}(\bm{q}) \leq \bm{0}\label{eq:theory:nlp:inequality-constraints},
\end{align}
\end{subequations}%
where 
$\bm{q}=[\bar{\bm{\xi}}_k^T,\bar{\bm{w}}_k^T,\dots,\bar{\bm{\xi}}_{k+N_T-1}^T,\bar{\bm{w}}_{k+N_T-1}^T,\bar{\bm{\xi}}_{k+N_T}^T]^T$, the cost function is approximated with the rectangle method
\begin{equation}
\phi(\bm{q}) = \sum_{j=k}^{k+N_T} \delta_t J_{pf}(t_j,\bar{\bm{\xi}}_j,\bar{\bm{w}}_j),
\end{equation}
 and
\begin{equation}
  \bm{f_e}(\bm{q}) =
  \begin{bmatrix}
    \bar{\bm{\xi}}_k - \bm{\xi}(t_k)\\
    \bar{\bm{\xi}}_{k+1} - F(\bar{\bm{\xi}}_k,\bar{\bm{w}}_k)\\
    \vdots\\
    \bar{\bm{\xi}}_{k+N_T} - F(\bar{\bm{\xi}}_{k+N_T-1},\bar{\bm{w}}_{k+N_T-1})
  \end{bmatrix}.
\end{equation}
The inequality constraints use
(\ref{eq:theory:pf-ocp:state-constraints})-(\ref{eq:theory:pf-ocp:constraint-function})
 enforced on $\bar{\bm{\xi}}_i$ for $i=k,\dots,k+N_t$.

\subsubsection{Collocation}
For the collocation method, with $d$ collocation points, we define $j=0,\dots,d$ Lagrange polynomials
\begin{equation}
  L_{j}(\tilde{\tau}) = \prod_{r=0,r\neq j}^{d}\frac{\tilde{\tau}-\theta_r}{\theta_j-\theta_r}
\end{equation}
where $\tilde{\tau}\in[0,1]$, $\theta_0$ is 0, and the other
$\theta_i$ are Legendre collocation points. The approximation of the
state trajectory between $t_k$ and $t_{k+1}$ is then
\begin{equation}
  \bar{\bm{\xi}}(\tau) = \sum_{j=0}^dL_{j}\left(\frac{\tau-t_k}{\delta_t}\right)\bar{\bm{\xi}}_{k,j}\text{, for }\tau\in[t_k,t_{k+1}]
\end{equation}
where $\bar{\bm{\xi}}_{k,j}$ are optimization variables included in
$\bm{q}$. Requiring the derivatives to be equal on the collocation
points, and the simultaneous constraint to hold, we have
\begin{align}
  &\bar{\bm{\xi}}_{k+1,0} = \sum_{j=0}^dL_{j}(1)\bar{\bm{\xi}}_{k,j}\\
  &\bm{f_{\xi}}(\bar{\bm{\xi}}_{k,j},\bar{\bm{w}}_k) - \frac{1}{\delta_t}\sum_{r=0}^d\dot{L}_{r}(\theta_j)=0\text{, for }j=1,\dots,d
\end{align}
where $L_{r}(1)$ and $\dot{L}_r(\theta_j)$ are independent of $t_k$
and are precomputed. This gives a similar structure to
(\ref{eq:theory:nlp}) with the cost function evaluated with
$\bar{\bm{\xi}}_{i,0}$ for $i=k,\dots,k+N_t$. We have chosen to
evaluate (\ref{eq:theory:pf-ocp:state-constraints}) at all states
$\bar{\bm{\xi}}_{k,j}$ and the nonlinear inequality constraints,
(\ref{eq:theory:pf-ocp:constraint-function}), at
$\bar{\bm{\xi}}_{k,0}$. This reduces the computational burden, but
allows the collocation points $\bar{\bm{\xi}}_{k,j}$ between $t_k$ and
$t_{k+1}$ to violate the nonlinear inequality constraints. The
optimization vector is of the form
\begin{align}
  \bm{q} = &[\bar{\bm{\xi}}_{k,0}^T,\bar{\bm{\xi}}_{k,1}^T,\dots,\bar{\bm{\xi}}_{k,d}^T,\bar{\bm{w}}_k^T,\nonumber\\
           &\dots,\bar{\bm{\xi}}_{k+N_T-1,d}^T,\bar{\bm{w}}_{k+N_T-1}^T,\nonumber\\
           &\bar{\bm{\xi}}_{k+N_T,0}^T,\dots,\bar{\bm{\xi}}_{k+N_T,d}^T]^T
\end{align}
and equality constraint function
\begin{align}
  \bm{f_e}(\bm{q})=
  \begin{bmatrix}
    \bar{\bm{\xi}}_{k,0} - \bm{\xi}(t_k)\\
    \bar{\bm{\xi}}_{k+1,0} - \sum_{r=0}^dL_{r}(1)\bar{\bm{\xi}}_{k,r}\\
    \bm{f_{\xi}}(\bar{\bm{\xi}}_{k,0},\bar{\bm{w}}_k) - \sum_{r=0}^d\dot{L}_{r}(\theta_0)\bar{\bm{\xi}}_{k,0}\\
    \vdots\\
    \bm{f_{\xi}}(\bar{\bm{\xi}}_{k,1},\bar{\bm{w}}_k) - \sum_{r=0}^d\dot{L}_{r}(\theta_1)\bar{\bm{\xi}}_{k,d}\\
    \vdots\\
    \comment{\bar{\bm{\xi}}_{k+N_T,0} - \sum_{r=0}^dL_{r}(1)\bar{\bm{\xi}}_{k,r}\\
    \bm{f_{\xi}}(\bar{\bm{\xi}}_{k+N_T-1,0},\bar{\bm{w}}_{k+N_T-1}) - \sum_{r=0}^d\dot{L}_{r}(\theta_0)\bar{\bm{\xi}}_{k+{N_T-1},0}\\
    \vdots\\
    \bm{f_{\xi}}(\bar{\bm{\xi}}_{k+N_T-1,d},\bar{\bm{w}}_{k+N_T-1}) - \sum_{r=0}^d\dot{L}_{r}(\theta_d)\bar{\bm{\xi}}_{k+{N_T-1},d}}
  \end{bmatrix}.
\end{align}

\subsubsection{Interior point solver}
Primal Interior point methods consider NLPs of the form
\begin{subequations}\label{eq:theory:nlp-ip}
  \begin{align}
    \min_{\tilde{\bm{q}}}&\quad{\phi}(\tilde{\bm{q}})\label{eq:theory:nlp-ip:objective} - \mu \sum_{i=0}^{\tilde{n}}\ln(\tilde{\bm{q}}_n)\\
    \text{s.t.:}&\nonumber\\
    &\bm{f_e}(\tilde{\bm{q}}) = \bm{0}\label{eq:theory:nlp-ip:equality-constraints}
  \end{align}
\end{subequations}
where $\tilde{\bm{q}}$ includes slack variables to make $\bm{h_e}$ an
equality constraint, and $\mu$ defines the steepness of the barrier
associated with the slack variables. For large values of $\mu$ the
$\ln$ term will dominate and the solution will tend to the middle of
the feasible region. As $\mu$ decreases, $\phi$ will dominate and the
solution will move towards the optimal solution. Solving for
decreasing $\mu$ will converge to the solution of
(\ref{eq:theory:nlp}). Interior point methods are difficult to
warm-start, as a too low $\mu$ may make certain slack variables
prematurely small and cause slow convergence. In the timing tests, we
have not used warm start as the initial states are random, but in the
MPC implementation we use the previous $\bar{\bm{\xi}}$ predictions as
an initial guess.

We will use the interior point solver IPOPT~\cite{Wachter2006}, a
primal-dual interior point solver, solving (\ref{eq:theory:nlp-ip})
using the primal-dual equations, see section 3.1 in
\cite{Wachter2006}. Interior point methods have consistent runtime
with respect to problem size, allowing us to potentially include more
states with little effect on runtime.

Convergence of the MPFC is ensured using terminal sets and penalties
as constructed in~\cite{Faulwasser2016conv} where an example is given
for the same system as ours with different parameters. In this article
we do not consider the terminal cost and penalty. Reasoning being that
most industrial machines have the accuracy to move to a set-point
sufficiently close to the path, allowing us focus on the run time.

\section{Simulation}
We consider a two-link manipulator. This can easily be
extended to 6 degrees-of-freedom, and results here are indicative of
the larger systems as interior point methods are consistent with
respect to the number of variables. The system was implemented using
Python and the CasADi framework~\cite{Andersson2013b}. CasADi allows
us to define symbolic expressions for the various equations in
(\ref{eq:theory:pf-ocp}), and evaluate the derivatives using
algorithmic differentiation, e.g. for RK4, which may be difficult to
do by hand. The framework supports IPOPT~\cite{Wachter2006}.

\subsection{System}\label{simulation:system}
\begin{figure}[h]
  \centering
  \includegraphics[width=0.49\columnwidth]{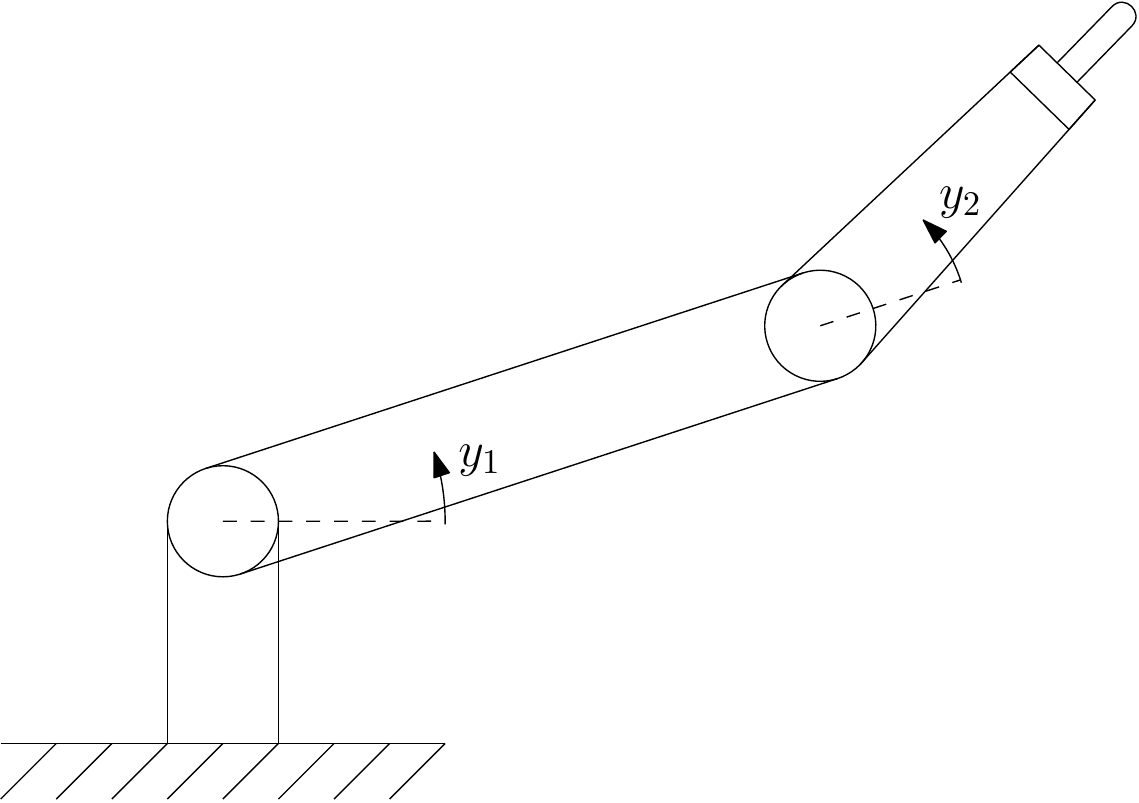}
  \caption{Two-link manipulator with 2 revolute joints.}
  \label{fig:two-link-manipulator}
\end{figure}
The robot has two links of length $L_1$ and $L_2$, with link masses
$M_1$ and $M_2$ at $L_{c1}$ and $L_{c2}$, and masses $m_1$ and $m_2$
at the joints. The point of interest is the tip of the end
effector. The system is described by
\begin{subequations}
  \begin{align}
    &\dot{\bm{y}} = \dot{\bm{x}}\\
    &\dot{\bm{x}} = \bm{M}(\bm{y})^{-1}\left(\bm{u} - \bm{C}(\bm{y},\bm{x})\bm{x} - \bm{G}(\bm{y})\right)
  \end{align}
\end{subequations}%
with 
\begin{equation}
  \label{eq:simulation-inertia-matrix}
  \bm{M}(\bm{y}) =
  \begin{bmatrix}
    a_1 + a_2\cos(y_2) & \frac{1}{2}a_2\cos(y_2) + a_3\\
    \frac{1}{2}a_2\cos(y_2)+a_3 & a_3
  \end{bmatrix}
\end{equation}
\begin{equation}
  \label{eq:simulation-coriolis-matrix}
  \bm{C}(\bm{y},\bm{x}) =
  \begin{bmatrix}
    -\frac{1}{2}a_2\sin(y_2)x_2 & -\frac{1}{2}\sin(y_2)(x_1+x_2)\\
    \frac{1}{2}a_2\sin(y_2)x_1 & 0
  \end{bmatrix}
\end{equation}
\begin{equation}
  \label{eq:simulation-gravity-vector}
  \bm{G}(\bm{y}) =
  \begin{bmatrix}
    g_1\cos(y_1)+g_2\cos(y_1+y_2)\\
    g_2\cos(y_1+y_2)
  \end{bmatrix},
\end{equation}%
where 
\begin{align}
  a_1 = I_1 + I_2 + m_1 L_{c1}^2 + m_2(L_1^2+L_2^2),\\
  a_2 = 2m_2L_{c2}L_1\text{, } \quad a_3 = m_2L_{c2}^2+I_2,\\
  g_1 = (L_{c1}(m_1+M_1) + L_1(m_2+M_2))g,\\
  g_2 = L_{c2}(m_2+M_2)g.
\end{align}

For brevity we give $a_1,a_2,a_3,g_1,g_2\in\mR$ in Table
\ref{tab:simulation:system-parameters}, for more information see
\cite{Astolfi2007}. The joint angles are defined as in
Fig.\ref{fig:two-link-manipulator}. The maximum torques are $30$ Nm,
the timesteps are $\delta_t = 0.01$ s, and if not otherwise specified
the horizon is $T = 0.20$ s.

\begin{table}[h]
  \centering
  \caption{System parameters}
  \begin{tabular}{c  c  c  c  c  c }
    \toprule
    Parameter & $a_1$ & $a_2$ & $a_3$ &$g_1$ &$g_2$ \\
    \midrule
    Value & 0.5578 &0.2263 &0.0785 & 17.0694 &4.3164\\
  \end{tabular}
  \label{tab:simulation:system-parameters}
\end{table}

\begin{table}[h]
  \centering
  \caption{MPC Parameters}
  \begin{tabular}{c c c c c c }
    \toprule
    Parameter & $\bm{Q}$ & $\bm{Q}_d$ & $\bm{R}$ & $q$ & $r$\\
    \midrule
    MPFC & $10^4\bm{I}_{2\times 2}$ & $10^1\bm{I}_{2\times 2}$ & $10^{-3}\bm{I}_{2\times 2}$ & $1$ & $ 10^{-3}$\\
    TT-MPC &  $10^4\bm{I}_{2\times 2}$ & $10^1\bm{I}_{2\times 2}$ & $10^{-3}\bm{I}_{2\times 2}$ & &\\
  \end{tabular}
  \label{tab:simulation:tuning}
\end{table}

In order to study obstacle avoidance, we include obstacles $o_i$ as bounding circles with known radius
$r_{o_i}$ and position $\bm{p}_{o_i}$. Their inequality equations are
\begin{equation}
  \bm{h}_{o_i} = \mnorm{\bm{f_p}(\bar{\bm{\xi}}(t)) - \bm{p}_{o_i}} - r_{o_i}^2 > 0.
\end{equation}
In actual applications a vision system would bound detected objects or
people by a circle that the point of interest is not to enter. When
present we consider two obstacles with: $r_{o_1}=0.02$m, at
$p_{o_1}=[0.55,0.75]^T$, and $r_{o_2}=0.04$m, at
$p_{o_2}=[0.4,0.4]^T$. 

The reference path is a circle of radius 0.2 with center at $[0.55, 0.55]^T$.

\subsection{Results}\label{simulation:results}
\subsubsection{Moving to origin}
\begin{figure}[h]
  \centering
  \includegraphics[width=0.89\columnwidth]{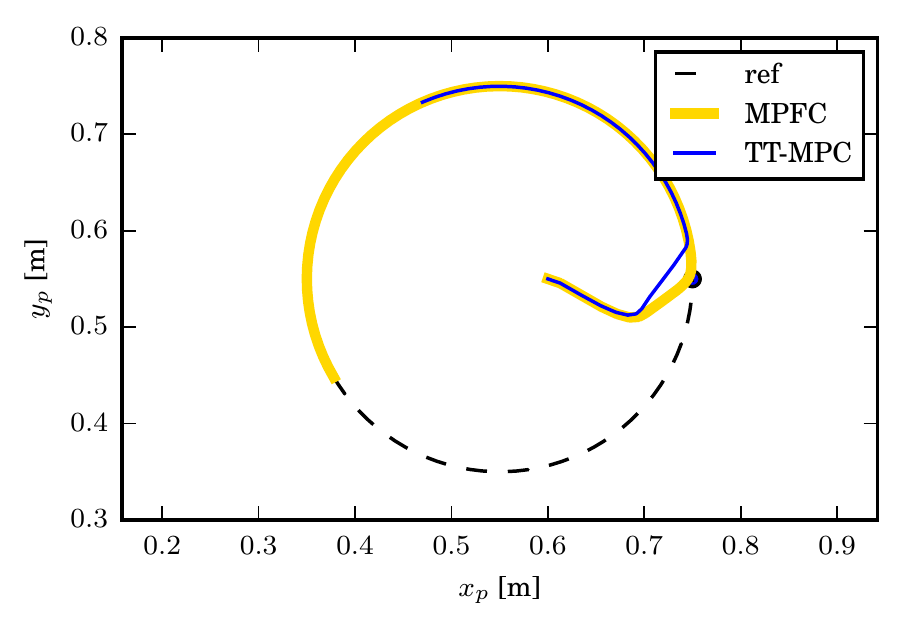}
  \caption{TT-MPC and MPFC moving from the same start point towards
    the path. The blue dot is the start of the reference path.}
  \label{fig:simulation:tt-vs-pf-cart}
\end{figure}

In
Fig.\ref{fig:simulation:tt-vs-pf-cart} we see the Cartesian paths of
the Runge-Kutta TT-MPC and MPFC. The black dot is $\varrho(0)$. For
this simulation we used maximum joint speeds of $0.5\pi$ rad/s to
exaggerate the differences. The set-point of the TT-MPC moves
gradually while the TT-MPC is approaching the path. The MPFC on the
other hand first approaches the origin of the path, then moves along
it. If $q$ is large compared to $\bm{Q}$, we will move along the path
faster than to the path. The MPFC has come further with no difference
in path deviation as $\dot{\bar{s}}$ is greater than the rate of $t$,
allowing it to move faster along the path.

\subsubsection{Obstacles}
\begin{figure}[h]
  \centering
  \includegraphics[width=0.89\columnwidth]{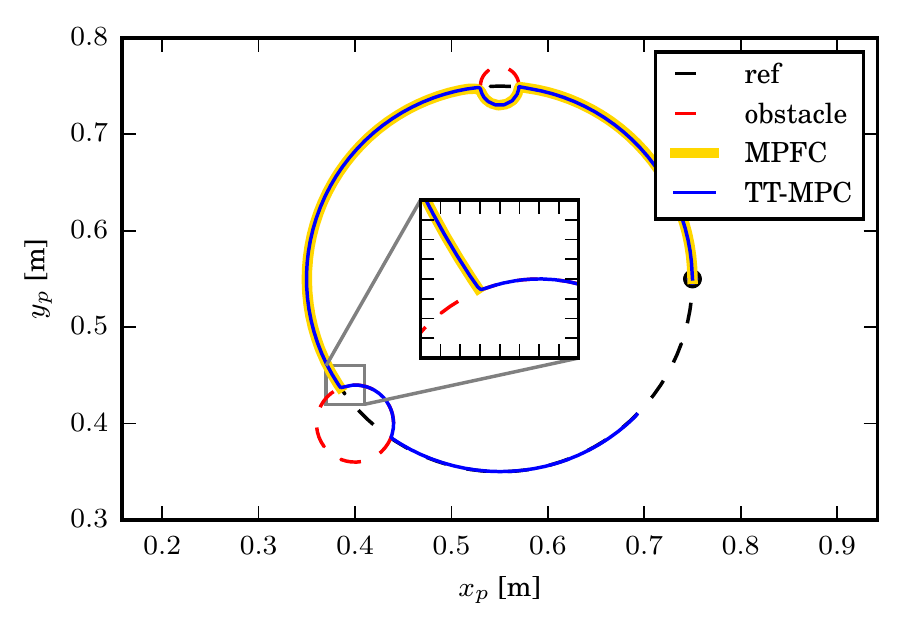}
  \caption{TT-MPC and MPFC initialized close to the path origin. The
    path is obstructed by a small and a large object. Both controllers
    pass the first object, but the MPFC does not pass the second.}
  \label{fig:simulation:tt-vs-pf-cart-obstacle}
\end{figure}

\begin{figure}[h]
  \centering
  \includegraphics[width=0.8\columnwidth]{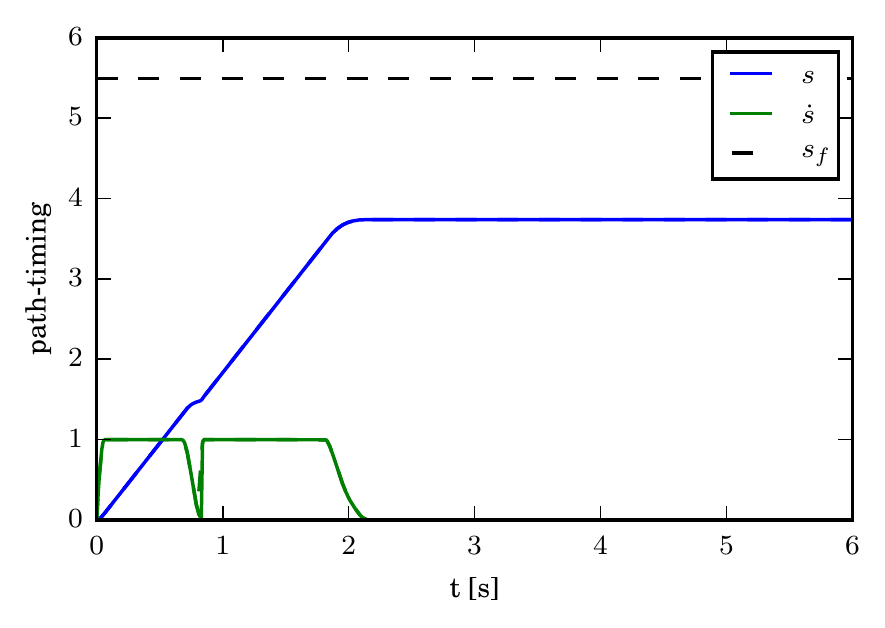}
  \caption{Timing parameter $s$, blue line, and $\dot{s}$, green line,
    when MPFC follows the obstructed path. First object encountered at
    $t\approx 0.9$s, second object encountered at $t\approx2$s.}
  \label{fig:simulation:z-obstructed}
\end{figure}

In Fig.\ref{fig:simulation:tt-vs-pf-cart-obstacle} two obstacles have
been placed in the path, and the speed constraints are removed. Both
MPFC and TT-MPC pass the first obstacle, but the MPFC stops at the
second. The second obstacle is too large, and the horizon is not long
enough to pass behind the obstacle. The MPFC will decrease
$\dot{\bar{s}}$ to zero, see $t\approx2$s in
Fig.\ref{fig:simulation:z-obstructed}, whereas the TT-MPC will have a
gradually increasing cost as the trajectory set-point moves forward
through the obstacle, forcing it around the object.

\begin{figure}[h]
  \centering
  \begin{subfigure}{\columnwidth}
    \centering
    \includegraphics[width=0.8\columnwidth]{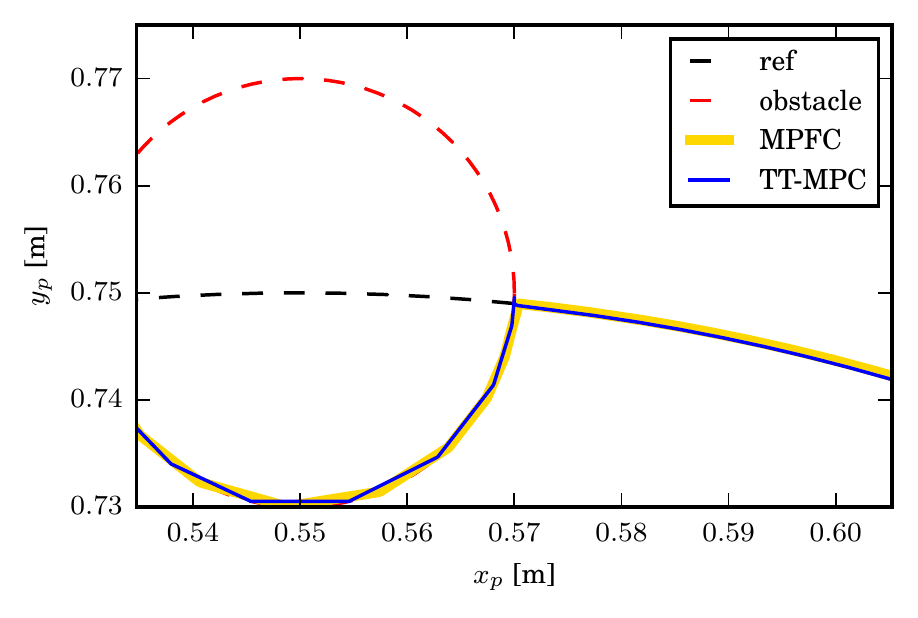}
    \caption{Collocation method}
  \end{subfigure}
  \begin{subfigure}{\columnwidth}
    \centering
    \includegraphics[width=0.8\columnwidth]{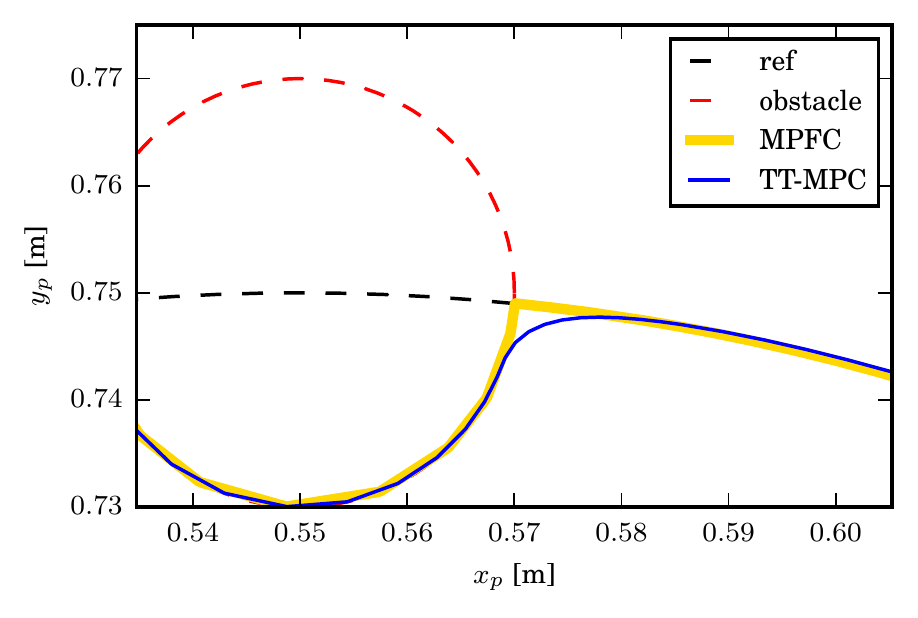}
    \caption{RK4}
  \end{subfigure}
  \caption{Difference between the two integration methods when the
    path is obstructed. Only evident in TT-MPC as the MPFC slows
    sufficiently down before the obstacle to not encounter the
    integration problems.}
  \label{fig:simulation:tt-vs-pf-cart-obstacle-zoom-both}
\end{figure}

When obstructed, there was a difference between the two integration
methods for the TT-MPC. We saw that the collocation method left the
TT-MPC path closer to the obstacles, see
Fig.\ref{fig:simulation:tt-vs-pf-cart-obstacle-zoom-both}. The MPFC
decreased $\dot{s}$ upon approaching the first object, at
$t\approx 0.9$s in Fig.\ref{fig:simulation:z-obstructed}, and was not
affected by integration differences. 

\subsubsection{Timing}
\begin{figure}[h]
  \centering
  \includegraphics[width=0.88\columnwidth]{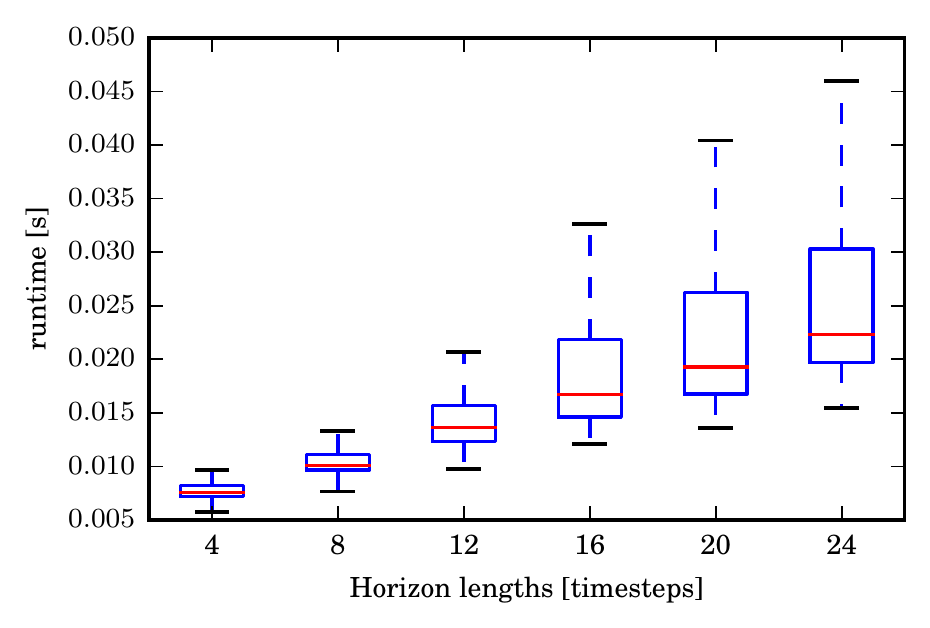}
  \caption{Boxplot of 2000 RK4 MPFC solver run times.}
  \label{fig:simulation:pf-drk-timing}
\end{figure}

\begin{figure}[h]
  \centering
  \includegraphics[width=0.88\columnwidth]{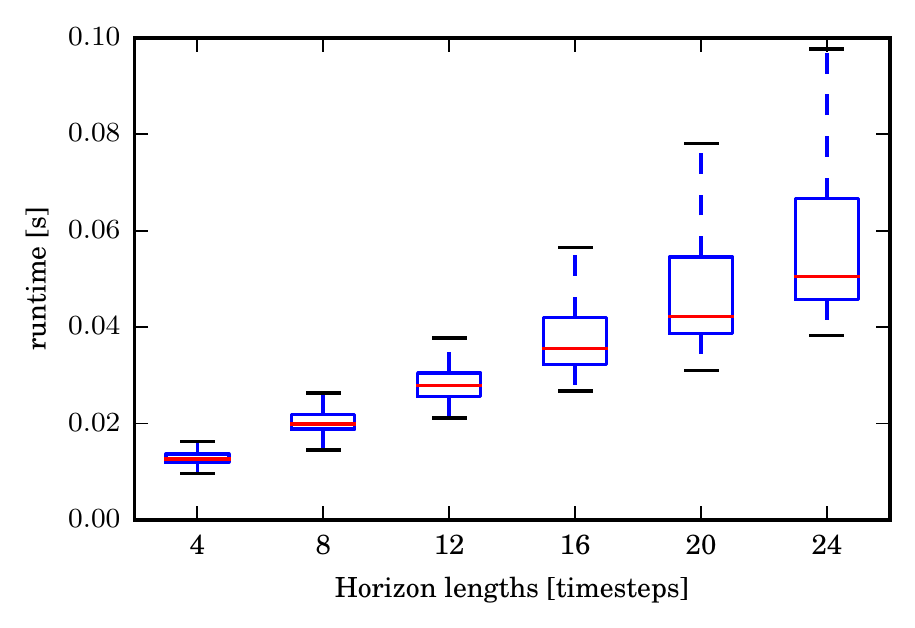}
  \caption{Boxplot of 2000 collocation MPFC solver run times.}
  \label{fig:simulation:pf-dc-timing}
\end{figure}

The simulations were performed on a Macbook Pro
with a 2.5 Ghz i7 CPU. Using the compilation feature of CasADi we can
create implementations that approach speeds needed for real-time
systems. To compare the timings of the two integration methods we have
performed a Monte-Carlo simulation of the MPFC with uniformly
distributed initial positions in the upper right quadrant of the
workspace. Box plot of the solver using RK4 for varying horizon
lengths is given in Fig.\ref{fig:simulation:pf-drk-timing}.  In
Fig.\ref{fig:simulation:pf-dc-timing} we give the same for the
collocation method.

CasADi gives timing statistics of the solver. Upon inspection it
appears that the collocation method has faster evaluation of the
constraint functions, Hessian of the problem Lagrangian, and generally
fewer iterations, but the increased optimization vector length makes
the solver slower. In Table \ref{tab:simulation:solvertimes} we give
typical timings of the solver. The cost function and cost gradient are
not included as they were the same and approximately 1 ms.


\begin{table}[h]
  \centering
  \caption{Typical timing statistics T = 0.6}\label{tab:simulation:solvertimes}
  \begin{tabular}{c c c c c c c}
    \toprule
    & Constr.\tablefootnote{The constraint function include both the inequality constraints and the ODE dynamics.} & $\nabla_{\bm{q}}$Constr. & $\bm{H}(\mathcal{L})$\tablefootnote{Hessian of the problem Lagrangian.}  & Iter. & Solver tot.\\
    \midrule
    MPFC-RK4 & 0.03 ms & 0.16 ms& 0.52 ms& 27& 0.045 s\\
    MPFC-Col  & 0.03 ms& 0.08 ms& 0.15 ms& 24&0.102 s\\
    TT-MPC-RK4 & 0.03 ms  & 0.23 ms& 0.66 ms& 14 & 0.039 s\\
    TT-MPC-Col & 0.03 ms  & 0.07 ms & 0.11 ms& 14  &0.046 s
  \end{tabular}
\end{table}

\section{Discussion and Future work}
The collocation method has a sparse structure in the equality
constraints, and relies on evaluation of $\bm{f_{\xi}}$, whereas the
RK4 requires evaluation of its algorithimic differentiation. This
gives the collocation method faster function evaluations, but it did
not appear to be sufficient to make the collocation method faster than
RK4. The solver itself took more time with the increased number of
states. For systems with complex dynamics the collocation method may
be necessary. TT-MPC had fewer states, simpler dynamics and was faster
on the whole, with the same observations as for the MPFC regarding
integration method.

The MPFC has the ability to stop along its path. For time-critical
systems this is not desired, but for robots in open environments it
can prove useful. It also suggests that when obstructed by an unknown
object, it may push against the object with a constant force. In
future experiments this will be investigated further. The TT-MPC will
observe a growing difference in the current position and the desired
position. With a known obstruction it will project the path onto the
constraint attempting to minimize the path error. Suddenly removing
the obstruction should result in the TT-MPC moving towards its current
set point as fast as possible. With an unknown obstruction the TT-MPC
may exert a gradually increasing force on the obstruction.

For the MPC to be real-time feasible, we require the solver to run
faster than the control interval used. In this article we have
considered a control interval of length $\delta_t=0.01$ s. For low
horizon lengths we are approaching such timing with the CasADi running
in Python. Future work will extend this framework for a 6
degrees-of-freedom robot with a 3D path. The low horizon length needed
to be able to achieve fast run time of the solver suggests that
terminal constraints may be required in the final system.

The obstructions considered in this article were static and known
apriori. Future work may include varying number of obstacles that
enter the robot workspace. 

\section{Conclusion}
The model predictive path-following controller gives rise to a set of
new design opportunities. Of most value for obstructed environments is
the fact that it may freely stop and resume along its path. The
question is whether a constraint ends the path, as the path-following
controller did, or whether the robot should move along the path
projected onto the constraint, as the trajectory tracking controller
did.

We also saw that the interior point method of IPOPT interfaced through
CasADi in Python, approached timings we would desire in a real-time
systems.

\section{Acknowledgements}\label{sec:acknowledgements}
The work reported in this paper was based on activities within centre
for research based innovation SFI Manufacturing in Norway, and is
partially funded by the Research Council of Norway under contract
number 237900. 

\bibliography{library.bib}{}

\begin{thebibliography}{10}
\providecommand{\url}[1]{#1}
\csname url@samestyle\endcsname
\providecommand{\newblock}{\relax}
\providecommand{\bibinfo}[2]{#2}
\providecommand{\BIBentrySTDinterwordspacing}{\spaceskip=0pt\relax}
\providecommand{\BIBentryALTinterwordstretchfactor}{4}
\providecommand{\BIBentryALTinterwordspacing}{\spaceskip=\fontdimen2\font plus
\BIBentryALTinterwordstretchfactor\fontdimen3\font minus
  \fontdimen4\font\relax}
\providecommand{\BIBforeignlanguage}[2]{{%
\expandafter\ifx\csname l@#1\endcsname\relax
\typeout{** WARNING: IEEEtran.bst: No hyphenation pattern has been}%
\typeout{** loaded for the language `#1'. Using the pattern for}%
\typeout{** the default language instead.}%
\else
\language=\csname l@#1\endcsname
\fi
#2}}
\providecommand{\BIBdecl}{\relax}
\BIBdecl

\bibitem{Verscheure2009}
D.~Verscheure, B.~Demeulenaere, J.~Swevers, J.~{De Schutter}, and M.~Diehl,
  ``{Time-optimal path tracking for robots: A convex optimization approach},''
  \emph{IEEE Transactions on Automatic Control}, vol.~54, no.~10, pp.
  2318--2327, 2009.

\bibitem{Debrouwere2013}
F.~Debrouwere, W.~{Van Loock}, G.~Pipeleers, M.~Diehl, J.~Swevers, and J.~{De
  Schutter}, ``{Convex time-optimal robot path following with Cartesian
  acceleration and inertial force and torque constraints},'' \emph{Proceedings
  of the Institution of Mechanical Engineers, Part I: Journal of Systems and
  Control Engineering}, vol. 227, no.~10, pp. 724--732, nov 2013.

\bibitem{Faulwasser2009}
T.~Faulwasser, B.~Kern, and R.~Findeisen, ``{Model predictive path-following
  for constrained nonlinear systems},'' \emph{Proceedings of the 48h IEEE
  Conference on Decision and Control (CDC) held jointly with 2009 28th Chinese
  Control Conference}, no.~3, pp. 8642--8647, 2009.

\bibitem{Faulwasser2016conv}
T.~Faulwasser and R.~Findeisen, ``{Nonlinear Model Predictive Control for
  Constrained Output Path Following},'' \emph{IEEE Transactions on Automatic
  Control}, vol. 9286, no.~c, pp. 1--1, 2016.

\bibitem{Faulwasser2016}
T.~Faulwasser, T.~Weber, P.~Zometa, and R.~Findeisen, ``{Implementation of
  Nonlinear Model Predictive Path-Following Control for an Industrial Robot},''
  \emph{IEEE Transactions on Control Systems Technology}, pp. 1--7, 2016.

\bibitem{Houska2011a}
B.~Houska, H.~J. Ferreau, and M.~Diehl, ``{ACADO Toolkit -- An Open Source
  Framework for Automatic Control and Dynamic Optimization},'' \emph{Optimal
  Control Applications and Methods}, vol.~32, no.~3, pp. 298--312, 2011.

\bibitem{Lam2010}
D.~Lam, C.~Manzie, and M.~Good, ``{Model predictive contouring control},'' in
  \emph{49th IEEE Conference on Decision and Control (CDC)}, vol.~86,
  no.~8.\hskip 1em plus 0.5em minus 0.4em\relax IEEE, dec 2010, pp. 6137--6142.

\bibitem{Lam2011}
------, ``{Application of model predictive contouring control to an X-Y
  table},'' \emph{IFAC Proceedings Volumes (IFAC-PapersOnline)}, vol.~18, no.
  PART 1, pp. 10\,325--10\,330, 2011.

\bibitem{Bock2014}
M.~B{\"{o}}ck and A.~Kugi, ``{Real-time nonlinear model predictive
  path-following control of a laboratory tower crane},'' \emph{IEEE
  Transactions on Control Systems Technology}, vol.~22, no.~4, pp. 1461--1473,
  2014.

\bibitem{Andersson2013b}
J.~Andersson, ``{A General-Purpose Software Framework for Dynamic
  Optimization},'' PhD thesis, Arenberg Doctoral School, KU Leuven, Belgium,
  2013.

\bibitem{modsim}
O.~Egeland and J.~T. Gravdahl, \emph{{Modeling and Simulation for Automatic
  Control}}.\hskip 1em plus 0.5em minus 0.4em\relax Trondheim: Marine
  Cybernetics, 2003.

\bibitem{Wachter2006}
A.~W{\"{a}}chter and L.~T. Biegler, \emph{{On the Implementation of Primal-Dual
  Interior Point Filter Line Search Algorithm for Large-Scale Nonlinear
  Programming}}, 2006, vol. 106, no.~1.

\bibitem{Astolfi2007}
A.~Astolfi, D.~Karagiannis, and R.~Ortega, \emph{{Nonlinear and Adaptive
  Control with Applications (Communications and Control Engineering)}}, 2007.

\end{thebibliography}
\bibliographystyle{IEEEtran}
\end{document}